\documentclass[a4paper, 10pt, conference]{ieeeconf}      

\IEEEoverridecommandlockouts                              

\usepackage{amsmath,amssymb,amsfonts}
\usepackage{algorithmic}
\usepackage{graphicx}
\usepackage{textcomp}
\usepackage{xcolor}
\usepackage{booktabs}   
\usepackage{makecell}   
\usepackage{multirow}
\usepackage{threeparttable}
\usepackage{adjustbox}
\usepackage{hhline}
\usepackage{array}
\usepackage{url}
\usepackage{mathtools}

\title{\LARGE \bf
Robust Localization, Mapping, and Navigation for Quadruped Robots
}

\author{Dyuman Aditya$^{1}$ , Junning Huang$^{2}$ , Nico Bohlinger$^{2}$ , Piotr Kicki$^{3,4}$ , Krzysztof Walas$^{3}$
,\\ Jan Peters$^{2,5,6}$ , Matteo Luperto$^{7}$ and Davide Tateo$^{2}$
\thanks{$^{1}$ {\'E}cole Centrale de Nantes, France}
\thanks{\tt\small dyuman.aditya@eleves.ec-nantes.fr}%
\thanks{$^{2}$ Technical University of Darmstadt, Germany}%
\thanks{$^{3}$ Poznan University of Technology, Poland} 
\thanks{$^{4}$ IDEAS NCBR, Warsaw, Poland}%
\thanks{$^{5}$ German Research Center for AI (DFKI), SAIROL}
\thanks{$^{6}$ Hessian.AI}
\thanks{$^{7}$ Universit{\`a} degli Studi di Milano, Italy}%
}

\DeclareMathOperator{\sign}{sign}

\begin{document}

\maketitle
\thispagestyle{empty}
\pagestyle{empty}

\begin{abstract}
Quadruped robots are currently a widespread platform for robotics research, thanks to powerful Reinforcement Learning controllers and the availability of cheap and robust commercial platforms.
However, to broaden the adoption of the technology in the real world, we require robust navigation stacks relying only on low-cost sensors such as depth cameras.
This paper presents a first step towards a robust localization, mapping, and navigation system for low-cost quadruped robots. In pursuit of this objective we combine contact-aided kinematic, visual-inertial odometry, and depth-stabilized vision, enhancing stability and accuracy of the system. 
Our results in simulation and two different real-world quadruped platforms show that our system can generate an accurate 2D map of the environment, robustly localize itself, and navigate autonomously.
Furthermore, we present in-depth ablation studies of the important components of the system and their impact on localization accuracy.
Videos, code, and additional experiments can be found on the project website.\footnote{\scriptsize\mbox{\url{https://sites.google.com/view/low-cost-quadruped-slam}}}
\end{abstract}

\section{INTRODUCTION}
The widespread availability of robust and reliable robotics platforms, and the major advances in locomotion controllers based on Reinforcement Learning~\cite{miki2022learning,rudin2022learning,margolis2022rapid,margolis2023walk,smith2023demonstrating,smith2024grow}, has enabled the deployment of quadruped robots in challenging environments, both outdoor and indoor~\cite{hoeller2021learning,tang2025deep}. Concurrently, there is an increasing interest in quadruped robots from the industry for surveillance, inspection, and search and rescue tasks, owing to their ability to navigate diverse settings—including stairs, cluttered spaces, and uneven terrain.

Unfortunately, practical industrial solutions typically rely on expensive robotic platforms equipped with high-precision sensors that enable robust autonomous navigation and mapping. In low-cost platforms, however, these sensors are not available, and most of the localization capabilities must rely on inexpensive depth image cameras or 2D LiDARs, resulting in subpar performance. These issues are further exacerbated when employing policies learned with Reinforcement Learning, which can achieve very high speeds and reactive behavior, at the expense of trajectory smoothness.
Such rapid trajectories may disrupt the SLAM pipeline, resulting in loss of localization, broken maps, and, consequently, the inability to navigate autonomously in the environment. These challenges are even more relevant when quadruped robots operate in indoor environments, thus in settings where fast movements change the robot's perceptions rapidly; in those environments, even the availability of expensive long-range LiDARS has a limited impact on performance, due to the proximity of obstacles to the robot's location.

\begin{figure}
\centering
\includegraphics[width=\linewidth]{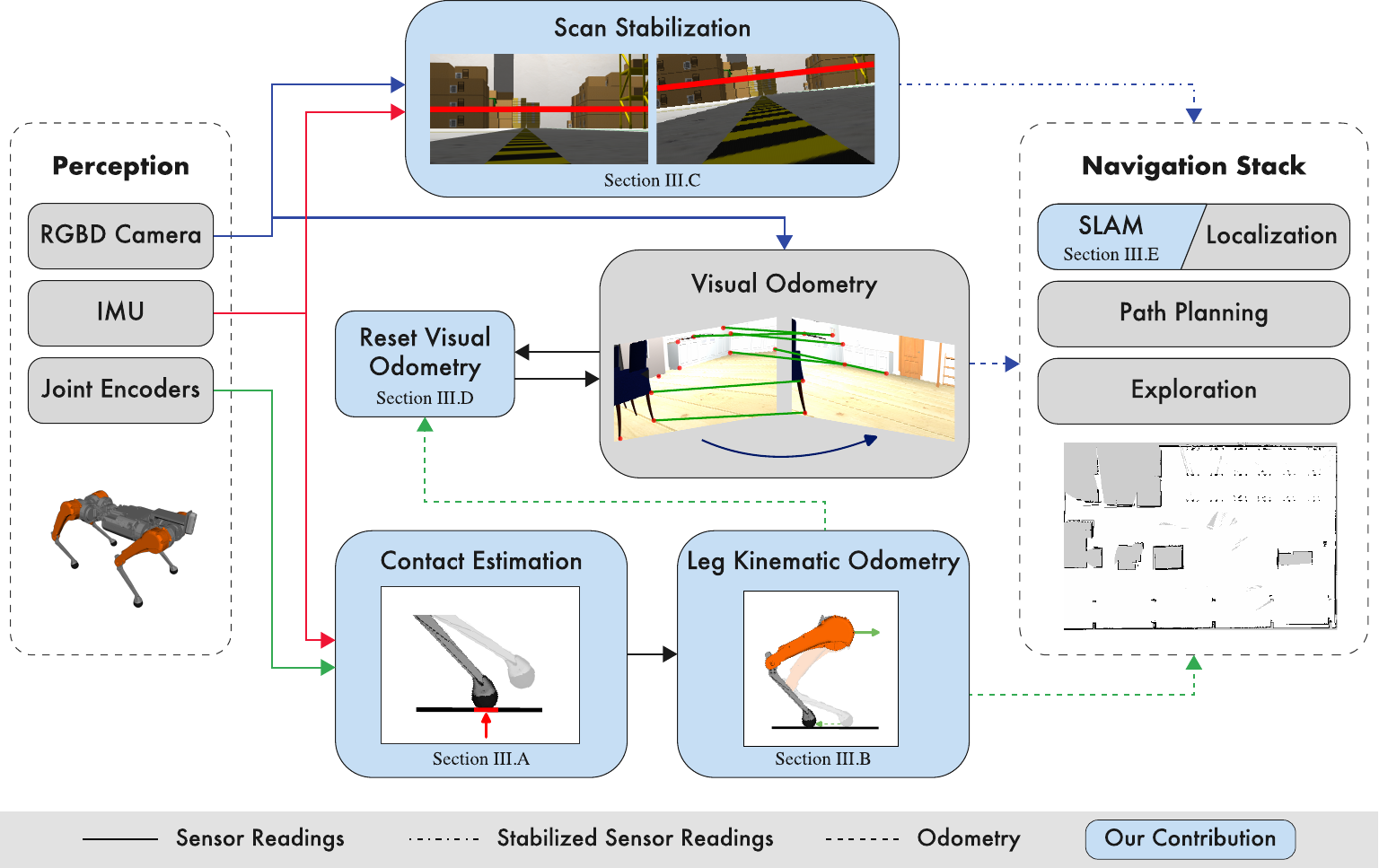}
\caption{Overall system design for robust navigation in low-cost quadruped robots. Our contributions are highlighted in light blue, different colors indicate different types of data.}
\label{fig:system_overview}
\end{figure}

In this paper, we focus on designing a simple and robust localization pipeline that enables low-cost quadruped robots to navigate autonomously in complex environments, thereby empowering these robots to tackle interesting real-world tasks. Our approach targets scenarios where the robot is equipped with a low-cost RGBD camera and an IMU. 
Due to the type and quality of sensors available for our target set-up, we restrict our localization and navigation pipeline to indoor settings. These environments, while being one of the target environments for quadruped robots, are particularly challenging as fast robot movements can cause abrupt changes in robot perception, thus causing localization and mapping issues. 
Moreover, we focus on localization, navigation, and mapping using 2D grid maps. Those maps are commonly used for path planning, localization, and path execution since widespread methods for 2D map-based navigation are still de facto the state-of-the-art for autonomous navigation~\cite{placed2023survey}, and since 2D maps can be combined with machine learning techniques and 3D and semantic maps~\cite{xiao2022motion}. 

In a 2D mapping setting, an occupancy map is generated using LiDAR scans or from depth data obtained by an RBGD camera~\cite{labbe2019rtab}. While setup is widely used in classical wheeled mobile platforms with stable performance, its implementation with quadruped robots is still challenging, and few robust solutions have been proposed yet. Due to rough, uneven, and slippery terrain, robot perception data are subject to jitters and noise, making 2D mapping and localization highly unreliable. Scans oscillate according to the terrain and the robot's own movement, especially when the movement involves sharp changes in roll and pitch angles.
Several works used 3D map representation and vision-based methods for SLAM with quadruped robots. Unfortunately, while 3D maps can provide a more semantically rich representation of the environment, they are still more computationally expensive to obtain and maintain. Moreover, when low-cost sensors are used, 3D maps are heavily affected by noise, making them impractical for navigation.

To enhance system autonomous navigation performance, we exploit joint encoder data to compute leg odometry~\cite{li2024robust}. This integration requires reliable contact information, typically obtained from contact sensors. However, these sensors are not always available or sufficiently accurate on legged platforms.  Recognizing that contact sensors are not always available or sufficiently accurate on legged robotic platforms, we also demonstrate how to estimate foot contacts using only joint torque measurements. Additionally, we incorporate a scan stabilization technique to mitigate the impact of rapid, dynamic movements on the SLAM pipeline, ensuring more robust and accurate localization.
The key contributions of this paper are i) a detailed description of the localization system, illustrating how it is possible improve localization and navigation performance integrating standard pipelines with specific problem-dependent solutions; our system involves a) an estimation of legged odometry and b) contact estimation, c) scan stabilization, and d) integration of leg-visual-inertial odometry to improve visual odometry. Then, we provide (ii) an extensive ablation study of the impact of each component of the localization system in simulation in several downstream tasks such as mapping, navigation, and exploration. Finally, we present (iii) an experimental real-world evaluation including a robot with an actuated spine.

\section{Related works}

Quadruped robots, similarly to other autonomous mobile robots, need localization, mapping, and autonomous navigation capabilities~\cite{zhou2024tightly}. While many different systems with such capabilities have been presented in literature, most of them rely on expensive robotic platforms, such as the Anymal and the Spot robots, mounting high-end, expensive long-distance sensors, thanks to their large form factor, allowing heavier payloads. An example of such systems can be found in ~\cite{lee2024learning} for the Anymal and in~\cite{raheema2024autonomous} for the Spot robot. Indeed, in~\cite{raheema2024autonomous}, the authors make a step towards low-end platforms by reducing the total amount of sensors used by the system. However, their setup is still relatively expensive compared with the setup considered in this paper. On top of this, we consider much more reactive and fast neural-based policies.
A similar setup is used in~\cite{zhou2024tightly}, where a long-distance 3D LiDAR and RGBD cameras are used for 3D SLAM in outdoor environments, while the recent work of Li et al.~\cite{li2025feature} presents a framework to perform 3D SLAM integrating features extracted from a long-distance LiDAR.

In~\cite{gupta2025fast}, the authors describe a framework designed to solve the task of autonomous exploration for object search, where a Spot robot is equipped with a Velodyne Puck LiDAR and an Azure Kinect RGBD camera. The robot plans its action using a 2D grid map by integrating a classifier to detect areas left to explore (labeled as \emph{non-map}) where the target object is located. Interestingly, even if a long-range expensive LiDAR is used, since acquiring mapping data for training models is difficult with quadruped robots due to their mechanical limitations, a remotely controlled wheeled cart with the same sensor equipment as the quadruped robot is used instead for data acquisition.

The authors of \cite{koval2022experimental} design an autonomy package that incorporates 3D LiDAR and IMU data for autonomous navigation of a Spot robot in GPS-constrained settings. The robot navigation is performed using a 2D occupancy map of the environment; to reduce the difficulty of building 2D maps of indoor environments with quadruped robots, a first run is made by manually moving the robot within the environment; the recorded data are later used to compute a 2D map, offline. In later runs, the map is used online for localization.

In~\cite{li2024robust}, the authors compute Leg Odometry (LO) by modeling the forward kinematics and fuse it with IMU data to correct the robot position among different keyframes. To stabilize the perception, dynamic and moving objects are filtered from camera data using computer vision methods. As a result of this, they provide a robust visual-slam method for quadruped robots in dynamic environments with visually challenging conditions.
Similarly, Kumar et al.~\cite{kumar2022periodic} improve visual-intertial SLAM by exploiting the periodic predictability in the motion of legged robots by performing visual SLAM separately on each portion of the gait circle, thus leveraging a model of how quadruped robots move.

The work of~\cite{cheng2024hybrid} investigates how the combination of different sources of odometry, such as laser odometry (LO) and laser-inertial odometry (LIO), reduces the uncertainty in 2D and 3D SLAM that is due to the high noise of legged odometry derived from kinematics in quadruped robots. To do so, they estimate LIO from 3D maps, using a long-range (70 meters) LiDAR and creating both a 2D grid map and a 3D point cloud map of the environment.

Bouman et al.~\cite{bouman2020autonomous} present a system for the large-scale and long-duration mapping and exploration of a GPS-denied indoor environment using a Spot Robot. The proposed setup is based on five custom RealSense RGBD cameras and a high-end long-range LiDAR. Different sources of odometry, such as kinematic-based odometry (KO) and LO, are integrated using an odometry multiplexer so that the noise in different sources of odometry can be compensated by integrating different inputs. Even if the robot is designed to work in a multi-floor environment, a 2D grid map is built and used for local and global path planning and navigation.

In~\cite{gan2025automated}, the authors present a system that uses a 2D map to improve the process of building a 3D map of the same environment. Using a long-range LiDAR sensor, a 2D map is used for navigation and to correct the expected localization within the 3D map, while the 3D map process uses LO and LIO derived from the 2D map. The decoupling of the process of 2D map-based navigation and 3D scene reconstruction allows focusing on the accuracy of the 3D reconstruction while having a robust 2D map.

Miller et al. \cite{miller2020mine} present a system for mapping large-scale GPS-denied environments using quadruped robots. The robot is equipped with a long-range LiDAR and a RealSense RGBD camera. 

Fahmi et al.~\cite{fahmi2022vital} present a terrain-aware locomotion framework for quadruped robots, where both proprioceptive and exteroceptive sensor readings are used to plan the feet contact location and the feet motion, and to plan the overall motion of the robot. This information is used to evaluate if the planned motion of the robot can cause a collision of a leg or place the robot's foot in an unstable position; the result of this model is used to improve motion control.

In~\cite{zhang2024learning}, the authors show how robot perception about reinforcement learning can be used to allow quadruped robots to navigate on risky terrains with dynamic motion. However, the system assumes perfect perception and uses a ground truth map for state estimation. A similar work, where a model of the environment is used for local navigation with sensor constraints is presented in~\cite{zhang2024resilient}.

Recently, Wasserman et al.~\cite{wasserman2024legolas} developed a method to estimate the robot odometry of a quadruped robot by integrating previous action and command velocities, IMUs, proprioceptive sensing, roll and pitch estimates, and the angles and velocities of each joint of the robot. These inputs are processed by a neural network tasked to provide a probabilistic estimate of the robot's translation and rotation.
Finally, several works proposed different approaches to integrate heterogeneous proprioceptive and exteroceptive sensor readings, including IMUs, joint encoders, contact sensors, force/torque measurements, camera, LiDAR, GPS, and motion capture, for state estimation in quadruped robots~\cite{kim2021legged,kim2022step,wisth2022vilens,yang2023cerberus}.

\section{SYSTEM DESIGN}
We present a system for robust localization, mapping, and navigation for quadruped robots in challenging environments by combining advances in visual-inertial odometry, contact-aided kinematic odometry, and an IMU-stabilized laser setup for mapping.
The overall system is described in Fig.~\ref{fig:system_overview}, where our contribution is highlighted in light blue. We introduce two key modules, one that performs legged-inertial odometry while also doing contact estimation (Section~\ref{sec:contact}-\ref{section:legged}), and another that generates stable scans from depth camera measurements and IMU attitude measurements (Section~\ref{section:SCAN}). We then integrate Visual-Inertial Odometry (VIO), Legged-Inertial Odometry (LIO), and laser-stabilized scans into existing 2D localization frameworks that we adapt to the particular settings of quadruped navigation (Section~\ref{sec:nav-setup}). Furthermore, we use the leg odometry information to restart the visual odometry system when it gets lost (Section~\ref{sec:integration}). This allows us to maintain a reasonable pose estimate even when the robot performs fast and/or abrupt movement, causing a loss of visual feature tracking (Section~\ref{sec:intSLAM}).

\subsection{Contact estimation}\label{sec:contact}
Low-cost quadruped robots are often not equipped with contact estimation sensors, as in the Dingo quadruped, and the RealAnt robot~\cite{boney2020realant}.
In the absence of external force sensing, we use a contact state detector based on the classic generalized momentum (GM) disturbance observer~\cite{bledt2018contact}. The GM-based observer is proposed as follows
\begin{equation*}\label{eq:GM_disturbance_observer}
    \begin{bmatrix}
        \mathbf{\dot{\hat{p}}} \\ \mathbf{\dot{\hat{f}}}
    \end{bmatrix}
    =
    \begin{bmatrix}
        \mathbf{0} & -\mathbf{J}^T \\
        \mathbf{0} & \mathbf{0} 
    \end{bmatrix}
    \begin{bmatrix}
        \mathbf{\hat{p}} \\ \mathbf{\hat{f}}
    \end{bmatrix} 
    + 
    \begin{bmatrix}
        \bar{\boldsymbol{\tau}} \\ \mathbf{0}
    \end{bmatrix}
    +
    \begin{bmatrix}
     \mathbf{L}_1 (\mathbf{p}-\mathbf{\hat{p}}) \\
     \mathbf{L}_2 (\mathbf{p}-\mathbf{\hat{p}}) 
    \end{bmatrix},
\end{equation*}
with $\mathbf{p}=\mathbf{M}(\mathbf{x})\mathbf{v}$, the measured generalized momentum, $\mathbf{M}$ denotes the joint space mass matrix, $\mathbf{\hat{p}}$ the estimated generalized momentum, and $\mathbf{\hat{f}}\in\mathbb{R}^{12}$ the measured contact force for four legs. $\bar{\boldsymbol{\tau}}=\boldsymbol{\tau}_{m}+\mathbf{C}^T\mathbf{v}-\mathbf{g}$, where $\boldsymbol{\tau}_{m}$ are the motor torques, $\mathbf{C}$ the Coriolis matrix, $\mathbf{g}$ the gravity, and $\mathbf{L}_1, \mathbf{L}_2$ the observer gains.

Since our goal is to detect contact states and not to reconstruct the full force vector, we impose two constraints on the observer: we estimate the force only on the $z$ axis, and clip the estimated force to be always positive.
Moreover, to alleviate the phase‐lag inherent in high-gain disturbance observers, we introduce a mixed‐mode observer that blends high-gain and sliding-mode strategies. This design requires only a global incremental affine bound on the force‐signal nonlinearities—a strictly weaker assumption than that of pure high-gain observers \cite{andrieu2021observer}. Specifically, the observer dynamics are as follows
\begin{equation*}\label{eq:mix_mode_disturbance_observer}
    \begin{bmatrix}
        \mathbf{\dot{\hat{{p}}}} \\ \mathbf{\dot{\hat{{f}}}}
    \end{bmatrix}
    =
    \begin{bmatrix}
        \mathbf{0} & -\mathbf{J}^T \\
        \mathbf{0} & \mathbf{0} 
    \end{bmatrix}
    \begin{bmatrix}
        \mathbf{\hat{p}} \\ \mathbf{\hat{f}}
    \end{bmatrix} 
    + 
    \begin{bmatrix}
        \bar{\boldsymbol{\tau}} \\ \mathbf{0}
    \end{bmatrix}
    +
    \begin{bmatrix}
     \mathbf{L} k_1(\mathbf{p}-\mathbf{\hat{p}}) \\
     \mathbf{L}^2 k_2(\mathbf{p}-\mathbf{\hat{p}}) 
    \end{bmatrix},
\end{equation*}
with $\mathbf{\hat{f}}=\
\begin{bmatrix}
\mathbf{0} & \mathbf{0} & \hat{f}_z
\end{bmatrix}$ the estimated contact force on only $z$ axis, and $k_1, k_2$ are defined as 
\begin{align*}
k_1(s) \coloneq q(s) & &
k_2(s) \coloneq \sign(s) + q(s)
\end{align*}
with $q(s) \coloneq \sign(s)|s|^{1/2}$ + s, $\sign(s)=1$ if $s>0$ and $\sign(s)=-1$ if $s<0$, $\mathbf{L}$ the gain parameter.
To obtain the contact states, we filter the estimated contact force to reduce oscillations during the steady state period, and we use feet-specific thresholds to determine the contact states.


\subsection{Legged Odometry as Least Squares}\label{section:legged}

Using leg odometry~\cite{camurri2025slamhandbook}, we can estimate the twist of the robot body, i.e., linear and angular velocities, as a function of the joint velocities of the legs in contact with the ground. This is given as an equation of the form
\begin{equation}\label{eq:body-vel-body}
    \mathbf{v}_b^l  = -\omega_b \times\mathbf{p}_l - \mathbf{v}_l , 
\end{equation}
with $\mathbf{v}_b^l$ the velocity of the body due to leg $l$, $\omega_b$ the angular velocity of the body, $\mathbf{p}_l$ and $\mathbf{v}_l$ the position and velocity (in body frame) respectively of the leg in contact. 
Notice that the equation above represents a 3DoF linear constraint on the 6DoF robot twists. Thus, we require at least two linearly independent constraints to completely identify the twist of the robot body, resulting in a system of linear constraints.
To ensure that we always have at least two linearly independent constraints and improve the condition number of the linear system, we add the angular velocity measured by the IMU sensor as an additional constraint. The resulting system of constraints can be written as
\begin{equation}
    \begin{bmatrix}
    \mathbf{A} \\
    \mathbf{A}_{\text{IMU}}
    \end{bmatrix}
    \mathbf{V}_b =
    \begin{bmatrix}
    \mathbf{b} \\
    \boldsymbol{\omega}_{\text{IMU}}
    \end{bmatrix}
\end{equation}
with the twist of the body $\mathbf{V}_b$, the IMU angular velocity measurements $\boldsymbol{\omega}_{\text{IMU}}$, and the matrices $\mathbf{A} \in \mathbb{R}^{3 \times 6}$,  $\mathbf{A}_{\text{IMU}}\in \mathbb{R}^{3N \times 6}$, and $\mathbf{b}\in \mathbb{R}^{3N}$ defined as
\begin{align*}
    \mathbf{A} = \begin{bmatrix}
    \mathbf{A}_1 \\
    \mathbf{A}_2 \\
    \vdots \\
    \mathbf{A}_N
    \end{bmatrix}&  & & 
     \mathbf{A}_{\text{IMU}} = \begin{bmatrix} 0_{3 \times 3} & \mathbf{I}_{3 \times 3} \end{bmatrix} &
    \mathbf{b} = \begin{bmatrix}
    \mathbf{b}_1 \\
    \mathbf{b}_2 \\
    \vdots \\
    \mathbf{b}_N
    \end{bmatrix}    
\end{align*}
where each block $\mathbf{A}_i$ and $\mathbf{b}_i$ represent a single constraint and can be written as
\begin{align*}
    \mathbf{A}_i &= \begin{bmatrix} \mathbf{I}_{3 \times 3} & -\mathbf{S}(\mathbf{p}_i) \end{bmatrix} \\
    \mathbf{b}_i &= -\mathbf{v}_i = -\mathbf{J}_i \dot{\pmb{\theta}}_i
\end{align*}
where $\mathbf{S}(\mathbf{p}_i)$ is the skew-symmetric matrix representation of the position of the i-th leg $\mathbf{p}_i$, while $\mathbf{J}_i$ and $\dot{\pmb{\theta}}_i$ are the Jacobian and the joint velocities of the i-th leg.

We can find a solution to this linear system as a least squares problem
\begin{equation}
    \mathbf{\hat{{V}}}_b = \arg\min_{\mathbf{V}_b} \left\| \mathbf{A} \mathbf{V}_b - \mathbf{b} \right\|^2. \label{eq:least_squares_odom}
\end{equation}

The problem described in~\eqref{eq:least_squares_odom} can be solved in closed form as follows
\begin{equation}\label{eq:closed-form-least-squares-solution}
    \mathbf{\hat{{V}}}_b = \left( \mathbf{A}^\top\mathbf{A} \right)^{-1} \mathbf{A}^\top \mathbf{b}
\end{equation}

\subsection{Scan Stabilization}\label{section:SCAN}
In this Section, we show how we can exploit the information from depth images and the robot's orientation to generate a stabilized scan. To this end, we use the onboard IMU to select a slice of the image data, making it invariant to jitters and significant changes to roll and pitch angles. In this way, we can obtain more stable 2D depth readings that can be used for  2D mapping, localization, and, ultimately, navigation. An example of the effect of the scan stabilization can be seen in Fig.~\ref{fig:system_overview} and an example of the result in Fig.~\ref{fig:map-scan-stabilization}.

We extract the slice of image data along a line of pixels whose slope is determined by the roll $r$ angle and whose intercept is determined by the pitch $p$ angle. A standard width of pixels above and below this line is combined to form the depth scan. This allows us to compensate for any rotation along the $x$ and $y$ axes. The slope of the reference line is easily computed directly from the roll angle. However, the intercept position is slightly more complex to compute as it depends on these two quantities.
We derive an expression for the intercept position in pixels $\Delta v$ for an object’s image in response to a pitch rotation $\theta$ of the camera.

Assuming that the z-axis of the camera frame lies along the camera's optical axis, a pitch rotation along the y-axis in the robot frame corresponds to a roll rotation along the x-axis in the camera frame.
Therefore, in the camera frame, the robot's pitch rotation is represented by the following roll rotation matrix:
\begin{equation*}
R_x(\theta) = \begin{bmatrix} 1 & 0 & 0 \\ 0 & \cos\theta & -\sin\theta \\ 0 & \sin\theta & \cos\theta \end{bmatrix}
\end{equation*}
For simplicity, we assume the initial coordinates of an object in the camera’s frame are \( (X, Y, Z) = (0, 0, Z) \), meaning the object is positioned along the optical axis at depth \( Z \).
After applying the rotation \( \theta \), the new coordinates of the object in the camera’s frame are:
\begin{equation*}
\begin{bmatrix} X_c \\ Y_c \\ Z_c \end{bmatrix} = R_x(\theta) \cdot \begin{bmatrix} 0 \\ 0 \\ Z \end{bmatrix} = \begin{bmatrix} 0 \\ -\sin\theta \cdot Z \\ \cos\theta \cdot Z \end{bmatrix}
\end{equation*}
Let $K$ be the camera’s intrinsic matrix
\begin{equation*}
    K = \begin{bmatrix} f_x & 0 & c_x \\ 0 & f_y & c_y \\ 0 & 0 & 1 \end{bmatrix}
\end{equation*}
where $f_x$ and $f_y$ are the focal lengths in pixels along the x- and y-axes, respectively, and $(c_x, c_y)$ represents the principal point in pixel coordinates.
Using the pinhole camera model, the 3D point in the camera frame projects onto the image plane as:
\begin{equation*}
\begin{bmatrix} u \\ v \\ 1 \end{bmatrix} = K \cdot \begin{bmatrix} X_c / Z_c \\ Y_c / Z_c \\ 1 \end{bmatrix}
\end{equation*}
Substituting $X_c / Z_c = 0$ and $Y_c / Z_c = -\tan\theta$, we obtain:
\begin{align*}
    u &= c_x + f_x \cdot 0 = c_x \\
    v &= c_y + f_y \cdot (-\tan\theta)
\end{align*}
The vertical shift in pixels $\Delta v = v - v_0 = -f_y \cdot \tan\theta$ due to the pitch angle $\theta$ is the difference between the new $v$ coordinate and the initial $v_0 = c_y$.  This gives us the intercept value where the reference line equation as a function of pixel coordinates becomes
\begin{equation}
    v = (u - x_c)\tan{(-r)} - f_y\tan(p),
    \label{eq:line-laser-stab}
\end{equation}
where $(u, v)$ represents the pixel coordinates, with the origin being on the top left, $x_c$ is the $x$ coordinate center of the image, and $f_y$ represents the camera's focal length. Combining the pixels above and below this line, we get a stabilized scan using only depth and IMU data.

\subsection{Integrating leg odometry with visual inerital odometry}\label{sec:integration}

When tracking is lost, we leverage the computed leg odometry to restart the visual odometry system.
Due to the abrupt and jerky nature of quadruped locomotion compared to wheeled systems, sudden pose changes are frequent. This can cause visual odometry systems to lose track when consecutive frames lack sufficient feature correspondence. Such systems are typically reset to the last computed pose; however, with fast-moving platforms like ours, this approach often lacks accuracy and degrades localization performance. To mitigate this issue, we integrate the last available pose with the body velocity computed by the leg odometry system over small time intervals, thereby generating a more accurate pose to reinitialize visual odometry. This mechanism is triggered when no correspondences are identified between two subsequent frames. 



\subsection{Integrating leg odometry with 2D SLAM}\label{sec:intSLAM}

Finally, we use leg odometry to impose velocity constraints on the 2D SLAM factor graph.
2D SLAM factor graphs are typically formed with pose constraints based on laser scans. Loop closures occur, and the factor graph is optimized when scans are matched with ones nearby.
However, when quadruped robots are used, due to noise and jitter that result from their fast motion, laser scans are often particularly difficult to match, thus affecting both the quality of mapping and localization and ultimately not allowing the robot to robustly navigate in challenging environments.
To address this issue, in addition to the scan pose constraints, we add a velocity constraint between consecutive poses, thereby restricting sudden position changes due to mismatched scans. This ensures that the optimized poses remain consistent with the robot's estimated body velocity between poses. This is integrated with a standard 2D SLAM pipeline, as explained in Section~\ref{sec:nav-setup}.

\section{Experimental Evaluation}
In this Section, we perform an in-depth evaluation of our localization system by ablating its components. We carry out our experiments using two different robots: the Unitree Go2 and the MAB Silver Badger. The key difference between these platforms is that the MAB Silver Badger features an additional degree of freedom due to its actuated spine. This results in slightly more erratic camera movements, which makes the localization even more challenging.  We also tested our system with three other robotic platforms, the Unitree Go1 and A1, and the MAB Honey Badger, with similar results. For sensing, we use an Intel RealSense D435i RGBD camera. Additional results can be found on our website. 


\begin{table*}[t]
  \centering
  \small
  \setlength{\tabcolsep}{7pt}
  \begin{adjustbox}{max width=\textwidth}
  \begin{threeparttable}
    \caption{AWS Small House \& Warehouse Environment Ablation Study for SB and Go2 Robots in 5 independent runs.}
    \label{tab:combined-results}
    \begin{tabular}{l l | c c c c | c c c c}
      \toprule
      \textbf{Robot} & \makecell{\textbf{Metric}\\ \textbf{(RMSE)}} & \multicolumn{4}{c|}{AWS Small Warehouse} & \multicolumn{4}{c}{AWS Small House} \\
      \cmidrule(lr){3-6} \cmidrule(lr){7-10}
       & & \textbf{Ours} & \makecell{\textbf{B+SS}} & \makecell{\textbf{B+LO}} & \textbf{B} & \textbf{Ours} & \makecell{\textbf{B+SS}} & \makecell{\textbf{B+LO}} & \textbf{B} \\
      \midrule
      \multirow{6}{*}{SB} 
         & ATE       & \textbf{0.34 $\pm$ 0.09} & 5.19 $\pm$ 2.24 & 0.44 $\pm$ 0.11 & 5.00 $\pm$ 2.38 & \textbf{0.42 $\pm$ 0.11} & 2.54 $\pm$ 2.48 & 0.54 $\pm$ 0.22 & 1.38 $\pm$ 0.85 \\
         & ARE       & \textbf{0.06 $\pm$ 0.02} & 0.75 $\pm$ 0.21 & 0.08 $\pm$ 0.03 & 0.93 $\pm$ 0.37 & \textbf{0.08 $\pm$ 0.05} & 0.52 $\pm$ 0.49 & 0.10 $\pm$ 0.04 & 0.35 $\pm$ 0.18 \\
         & APE       & \textbf{0.35 $\pm$ 0.02} & 0.45 $\pm$ 0.11 & 7.90 $\pm$ 0.83 & 5.08 $\pm$ 2.41 & \textbf{0.43 $\pm$ 0.12} & 0.55 $\pm$ 0.22 & 2.59 $\pm$ 2.53 & 1.42 $\pm$ 0.86 \\
         & RPE (2m)  & \textbf{0.18 $\pm$ 0.01} & 0.18 $\pm$ 0.04 & 0.56 $\pm$ 0.04 & 0.51 $\pm$ 0.04 & \textbf{0.18 $\pm$ 0.02} & 0.21 $\pm$ 0.03 & 0.37 $\pm$ 0.04 & 0.27 $\pm$ 0.02 \\
         & RPE (5m)  & \textbf{0.26 $\pm$ 0.02} & 0.28 $\pm$ 0.11 & 1.15 $\pm$ 0.08 & 1.14 $\pm$ 0.16 & \textbf{0.30 $\pm$ 0.08} & 0.40 $\pm$ 0.11 & 0.70 $\pm$ 0.08 & 0.55 $\pm$ 0.04 \\
         & RPE (10m) & \textbf{0.31 $\pm$ 0.01} & 0.37 $\pm$ 0.19 & 2.09 $\pm$ 0.16 & 2.20 $\pm$ 0.46 & \textbf{0.45 $\pm$ 0.14} & 0.64 $\pm$ 0.23 & 1.08 $\pm$ 0.28 & 0.94 $\pm$ 0.10 \\
         
      \midrule
      \multirow{6}{*}{Go2} 
         & ATE       & \textbf{0.44 $\pm$ 0.07} & 0.85 $\pm$ 0.57 & 0.56 $\pm$ 0.20 & 0.53 $\pm$ 0.13 & \textbf{0.43 $\pm$ 0.22} & 0.43 $\pm$ 0.25 & 0.97 $\pm$ 0.63 & 0.93 $\pm$ 1.06 \\
         & ARE       & \textbf{0.08 $\pm$ 0.02} & 0.12 $\pm$ 0.05 & 0.10 $\pm$ 0.03 & 0.08 $\pm$ 0.04 & \textbf{0.11 $\pm$ 0.06} & 0.12 $\pm$ 0.06 & 0.21 $\pm$ 0.18 & 0.24 $\pm$ 0.29 \\
         & APE       & \textbf{0.45 $\pm$ 0.07} & 0.57 $\pm$ 0.19 & 0.86 $\pm$ 0.57 & 0.54 $\pm$ 0.13 & \textbf{0.45 $\pm$ 0.23} & 0.99 $\pm$ 0.65 & 0.44 $\pm$ 0.25 & 0.96 $\pm$ 1.10 \\
         & RPE (2m)  & \textbf{0.19 $\pm$ 0.03} & 0.23 $\pm$ 0.03 & 0.22 $\pm$ 0.06 & 0.23 $\pm$ 0.05 & \textbf{0.19 $\pm$ 0.04} & 0.23 $\pm$ 0.03 & 0.20 $\pm$ 0.05 & 0.23 $\pm$ 0.09 \\
         & RPE (5m)  & \textbf{0.27 $\pm$ 0.03} & 0.32 $\pm$ 0.05 & 0.38 $\pm$ 0.16 & 0.32 $\pm$ 0.07 & \textbf{0.25 $\pm$ 0.07} & 0.45 $\pm$ 0.18 & 0.26 $\pm$ 0.08 & 0.36 $\pm$ 0.16 \\
         & RPE (10m) & \textbf{0.37 $\pm$ 0.07} & 0.41 $\pm$ 0.11 & 0.54 $\pm$ 0.24 & 0.42 $\pm$ 0.10 & \textbf{0.36 $\pm$ 0.12} & 0.80 $\pm$ 0.38 & 0.40 $\pm$ 0.16 & 0.58 $\pm$ 0.38 \\
         
      \bottomrule
    \end{tabular}
    \begin{tablenotes}
      \footnotesize
      \item \textit{Acronyms:} SB = Silver Badger, Go2 = Unitree Go2, B+LO = Baseline + Leg Odometry, B+SS = Baseline + Scan Stabilization, B = Baseline.
    \end{tablenotes}
  \end{threeparttable}
  \end{adjustbox}
\end{table*}

In our experiments, we first show that our system can accurately estimate the state of the robot in simulation and real-world systems. While we do not strictly focus on accurate state estimation, this measure is useful for estimating the precision of map reconstruction. Indeed, a sufficiently good map is fundamental to perform autonomous navigation and exploration. Then, we present our results for autonomous navigation in known maps, showing that the system can navigate autonomously and robustly despite the low-cost sensors, without causing map corruption. We show that the system can be combined with an autonomous exploration algorithm. Finally, we report results made using two real-world settings where performance is measured using an OptiTrack system (Section~\ref{sec:realrobots}). For all experiments, we use the evaluation pipeline from~\cite{grupp2017evo}.

\subsection{Experimental Setting}\label{sec:exp:setting}
We develop our system in ROS2, and we integrate it with the ROS2 navigation stack. We provide both results obtained in realistic simulations, using Gazebo, and in the real world with quadrupeds using RL-based control policies~\cite{rudin2022learning}.

To compute precise metrics and rigorously evaluate our system, we conduct experiments in two large-scale simulated scenarios: the AWS small warehouse (a medium-scale furnished simulated warehouse) and the AWS small house environment (a three-room furnished simulated apartment).  To demonstrate the effectiveness of our approach, we perform an ablation study comparing our full method with configurations that exclude scan stabilization or legged odometry, as well as a baseline that relies solely on visual-inertial odometry combined with 2D SLAM. For simulated environments, we do not use the contact estimation module to highlight the contribution of the other components, and we use the ground-truth contact instead. We test the full model, including contact estimation, in the more challenging experiments made with a real robot (Section~\ref{sec:realrobots}). Full details can be found on our website.

Given that ground truth positions are available in simulation, we evaluate localization accuracy using the following six metrics~\cite{6385773}: \textbf{ATE} (Absolute Translation Error) and \textbf{ARE} (Absolute Rotational Error) measure, respectively, the Euclidean distance and the average angular distance between the estimated and ground truth positions over the entire trajectory. These two metrics are combined into the \textbf{APE} (Absolute Pose Error), which assesses the overall pose discrepancy in pose. Finally, the \textbf{RPE} (Relative Pose Error) evaluates the relative pose error over fixed intervals; we consider three values (2m), (5m), and (10m) representing different translations between poses.

\begin{figure}[b]
    \centering
    \includegraphics[height=0.49\columnwidth]{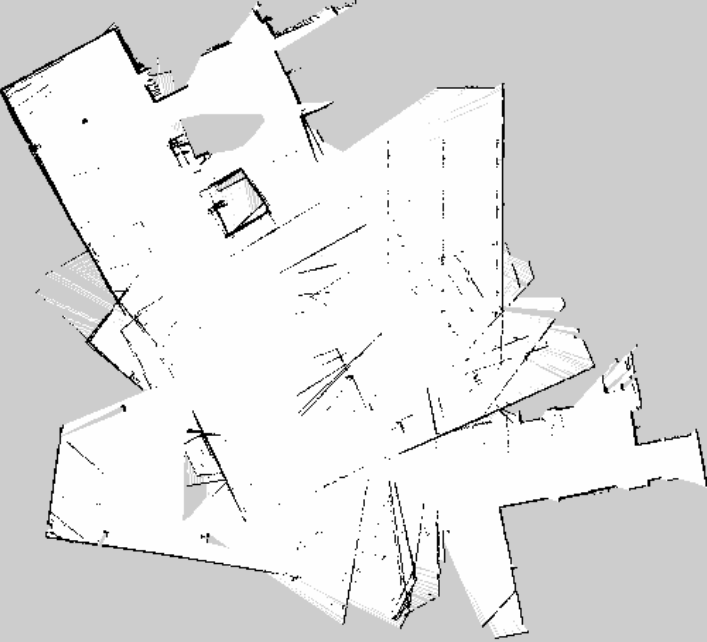}
    \includegraphics[height=0.49\columnwidth]{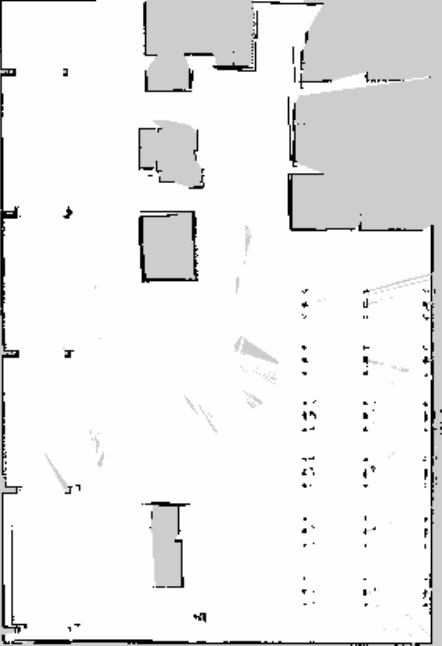}
    \caption{Effects of scan stabilization in the map generation. Left: no scan stabilization. Right: with scan stabilization.}
    \label{fig:map-scan-stabilization}
\end{figure}

\subsection{Integrated Navigation Stack}\label{sec:nav-setup}
Our navigation system is constructed entirely from ROS-compatible components and is integrated with the ROS2 navigation stack. We employ RTAB-Map~\cite{labbe2019rtab} as the front-end for visual-inertial odometry (VIO), utilizing RGB-D imagery in conjunction with IMU data to estimate motion in real time. This front-end provides robust pose estimates in visually structured environments.

For the SLAM back-end, we integrate our pipeline with the ROS \texttt{slam\_toolbox}~\cite{macenski2021slam}, which provides efficient pose graph optimization and map management capabilities. This modular combination allows our system to benefit from the real-time performance of RTAB-Map and the global consistency afforded by the graph-based SLAM framework.

For autonomous navigation on pre-mapped environments, we utilize the Adaptive Monte Carlo Localization (AMCL) in tandem with RTAB-Map’s visual odometry to ensure accurate and drift-resilient localization. This hybrid localization strategy enhances robustness, particularly in visually repetitive indoor settings, where AMCL alone might struggle.

\begin{figure}[b]
    \vspace{-1em}
    \centering
    \includegraphics[height=0.41\columnwidth]{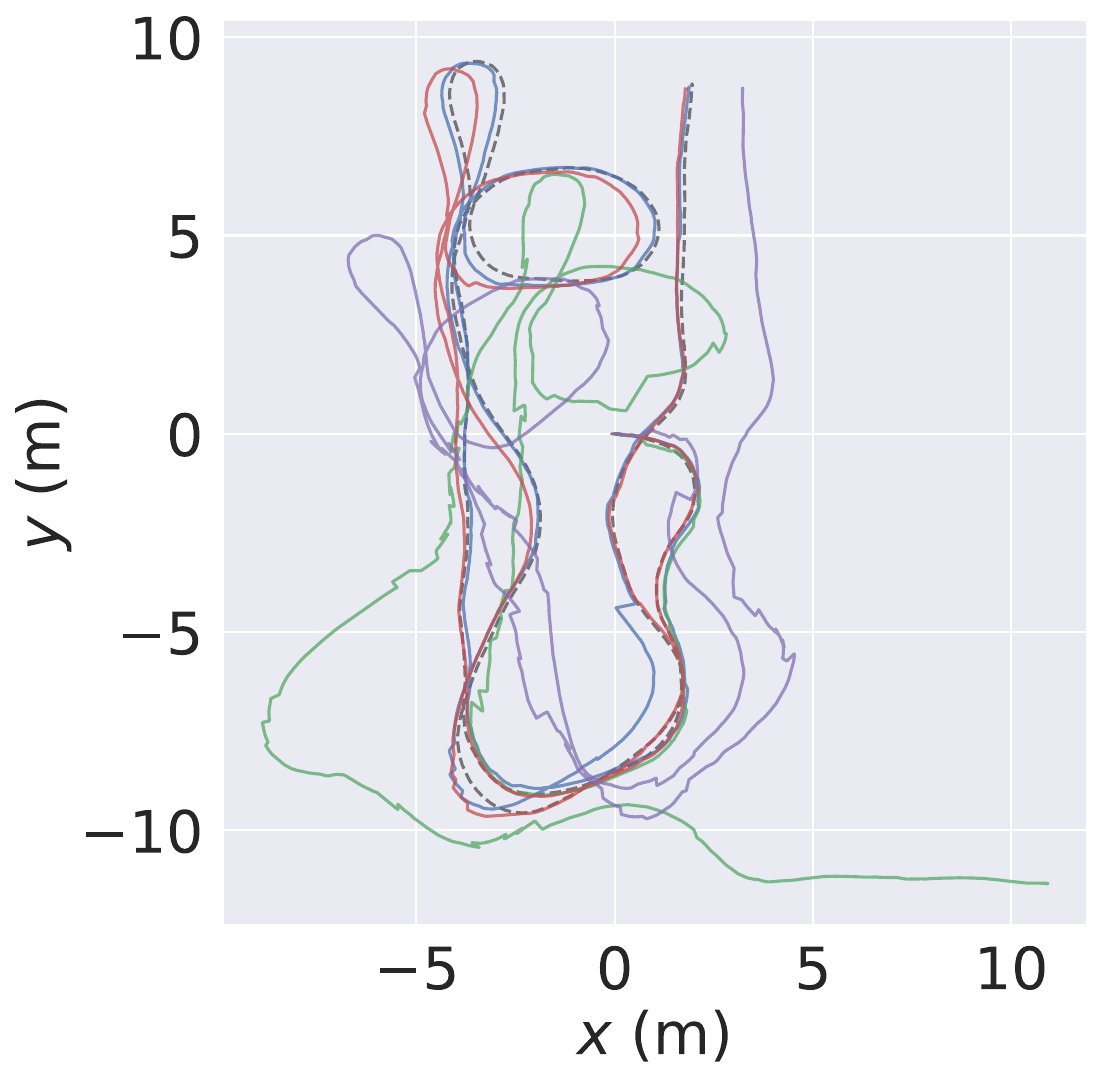}
    \includegraphics[height=0.41\columnwidth]{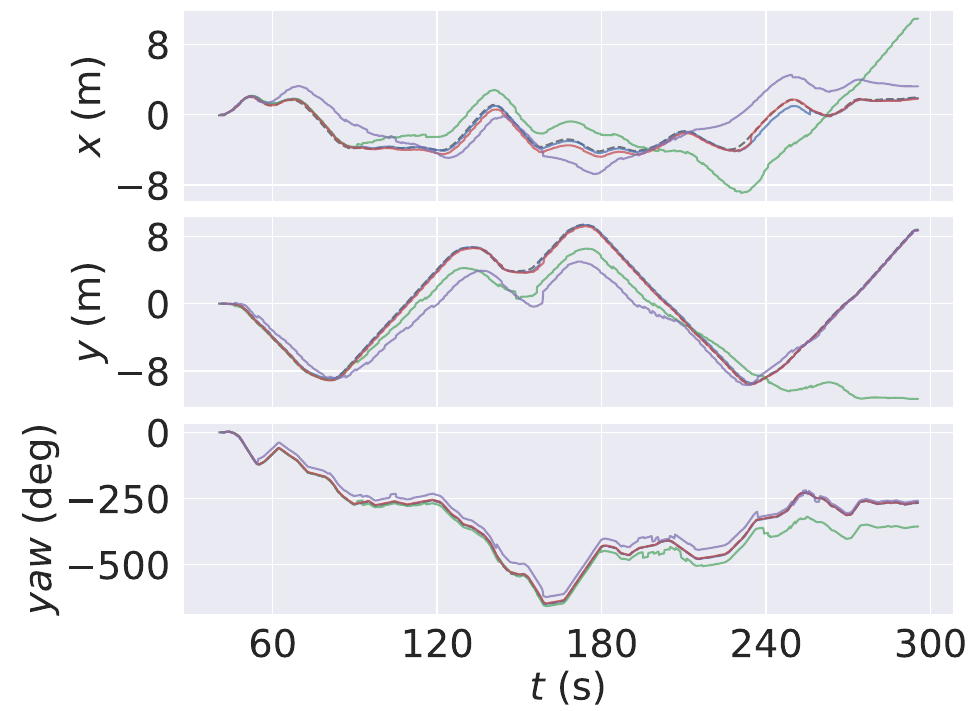}
    \includegraphics[width=\columnwidth]{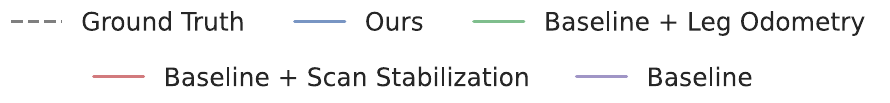}
    \caption{Estimated trajectories of the Silver badger robot in the small warehouse environment}
    \label{fig:warehouse_sb_traj}
\end{figure}

\subsection{Mapping Results}\label{sec:exp:results}
Table~\ref{tab:combined-results} presents the results of our ablation study, with metrics averaged over five runs per environment and robot. The baseline configuration (B) relies solely on visual-inertial odometry and laser-based mapping without additional enhancements. The system variants include B (Baseline), B+LO (Baseline with Leg Odometry), and B+SS (Baseline with Scan Stabilization). Each subsequent variant isolates the contribution of a specific component to assess its impact. The results clearly demonstrate that the full system outperforms the alternatives and maintains consistency across all the possible metrics. 
Notably, scan stabilization proves to be the most critical component--a conclusion further supported by the qualitative results in Fig.~\ref{fig:map-scan-stabilization}. This component is essential for generating accurate maps, enabling effective loop closure, and maintaining localization even during rapid movements that may challenge the VIO pipeline.



\begin{table}[b]
  \vspace{-1em}
  \centering
  \small
  \setlength{\tabcolsep}{7pt}
  \begin{adjustbox}{max width=\columnwidth}
  \begin{threeparttable}
    \caption{Navigation success rate results on AWS House and AWS Warehouse for SB and Go2 Robots.}
    \label{tab:navigation-table}
    \begin{tabular}{l | cc | cc}
      \toprule
      \textbf{Method} & \multicolumn{2}{c|}{AWS House} & \multicolumn{2}{c}{AWS Warehouse} \\
      \cmidrule(lr){2-3} \cmidrule(lr){4-5}
                     & \textbf{SB}  & \textbf{Go2}  & \textbf{SB}  & \textbf{Go2}  \\
      \midrule
      Baseline       & 0\%          & 80\%         & 60\%         & 60\%         \\
      Ours           & \textbf{100\%}        & \textbf{100\%}        & \textbf{100\%}        & \textbf{100\%}        \\
      \bottomrule
    \end{tabular}
    \begin{tablenotes}
      \footnotesize
      \item \textit{Acronyms:} SB = Silver Badger, Go2 = Unitree Go2
    \end{tablenotes}
  \end{threeparttable}
  \end{adjustbox}
\end{table}


Our experiments further reveal that incorporating leg odometry into the pipeline enhances localization accuracy, yielding slightly better maps. However, when leg odometry is used without laser stabilization, performance may occasionally degrade relative to the baseline. We attribute this to inaccuracies in scan alignment within the 2D SLAM system, which result in erroneous pose refinements. When combined with additional constraints from the leg odometry system, these inaccuracies further deteriorate overall performance. 

Fig.~\ref{fig:warehouse_sb_traj} illustrates that the full system accurately estimates trajectories that closely match the ground truth, underscoring the crucial role of scan stabilization in producing high-quality maps and enabling effective loop closure.
More detailed results, including per-run localization performance, are available on our project website.


\begin{table*}[t]
  \centering
  \small
  \setlength{\tabcolsep}{7pt}
  \begin{adjustbox}{max width=\textwidth}
  \begin{threeparttable}
    \caption{Real World Environment Ablation Study for the SB Robot in 5 independent runs.}
    \label{tab:combined-results-real}
    \begin{tabular}{l l | c c c c | c c c c}
      \toprule
      \textbf{Robot} & \makecell{\textbf{Metric}\\ \textbf{(RMSE)}} & \multicolumn{4}{c|}{Environment 1 - IAS Lab} & \multicolumn{4}{c}{Environment 2 - IRIM Lab} \\
      \cmidrule(lr){3-6} \cmidrule(lr){7-10}
       & & \textbf{Ours} & \makecell{\textbf{B+SS}} & \makecell{\textbf{B+LO}} & \textbf{B} & \textbf{Ours} & \makecell{\textbf{B+SS}} & \makecell{\textbf{B+LO}} & \textbf{B} \\
      \midrule
      \multirow{3}{*}{SB (real)} 
         & ATE         & \textbf{0.22 $\pm$ 0.36} & 1.91 $\pm$ 0.33 & 2.34 $\pm$ 0.43 & 2.51 $\pm$ 0.50 & \textbf{0.65 $\pm$ 0.28} & 1.68 $\pm$ 1.27 & 1.74 $\pm$ 0.77 & 1.50 $\pm$ 0.89 \\
         & ARE         & \textbf{0.33 $\pm$ 0.21} & 1.38 $\pm$ 0.33 & 1.65 $\pm$ 0.23 & 2.07 $\pm$ 0.17 & \textbf{0.48 $\pm$ 0.50} & 0.54 $\pm$ 0.39 & 0.78 $\pm$ 0.48 & 0.60 $\pm$ 0.38 \\
         & APE         & \textbf{0.64 $\pm$ 0.40} & 2.38 $\pm$ 0.28 & 2.83 $\pm$ 0.40 & 3.27 $\pm$ 0.37 & \textbf{0.84 $\pm$ 0.52} & 1.74 $\pm$ 1.23 & 1.99 $\pm$ 0.61 & 1.65 $\pm$ 0.90 \\
      \bottomrule
    \end{tabular}
    \begin{tablenotes}
      \footnotesize
      \item \textit{Acronyms:} SB = Silver Badger, B+LO = Baseline + Leg Odometry, B+SS = Baseline + Scan Stabilization, B = Baseline.
    \end{tablenotes}
  \end{threeparttable}
  \end{adjustbox}
  \vspace{-1em}
\end{table*}

\subsection{Robot autonomy}
After assessing our system’s map reconstruction capabilities, we proceeded to evaluate the robot’s autonomy and the robustness of its navigation stack.
First, we examined the navigation performance within a pre-established map across two environments using both the Silver Badger and Go2 platforms and using the previously described navigation setup (Section~\ref{sec:nav-setup}). The maps were generated before testing by teleoperating the robot through each environment, using our mapping system described in Section~\ref{sec:exp:results}. In this experiment, the robot is sequentially commanded to navigate to five predefined poses, and we measure its success rate in reaching each target. The results, presented in Table~\ref{tab:navigation-table}, clearly indicate that our system outperforms the baseline—achieving a higher success rate in reaching the commanded pose without losing track of its position w.r.t. the map. Scan stabilization plays a crucial role because the localization uses scans to localize itself. Even though the baseline method can navigate to a few poses successfully, the robot struggles to do so, and the navigation system has to recover several times. Videos of the results can be viewed on our project website.

\begin{figure}[b]
    \centering
    \includegraphics[angle=90,height=0.32\columnwidth]{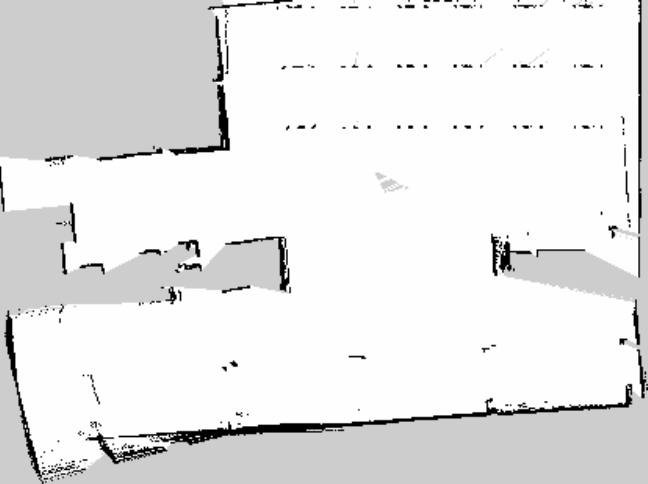}%
    \hspace{0.8em}%
    \includegraphics[angle=90,height=0.32\columnwidth]{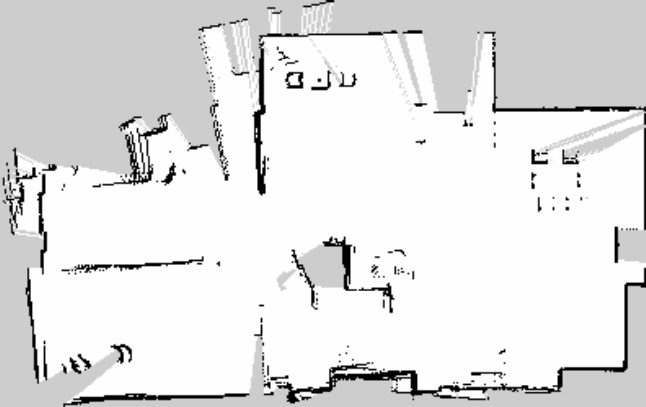}
    \caption{Autonomous exploration mapping results.}
    \label{fig:exploration-maps}
\end{figure}

Second, we evaluated the robot’s capability to solve the widely-known task of autonomous exploration for map building by using a frontier-based exploration strategy~\cite{YamauchiFrontier} using a nearest-frontier method. In this scenario, we use our mapping setup in conjunction with the navigation setup (Sections~\ref{sec:exp:results} \&~\ref{sec:nav-setup}) to autonomously map the environment.
The exploration results, available on our project website and illustrated in Fig~\ref{fig:exploration-maps}, demonstrate that the system can successfully navigate through diverse simulated environments. It covers over 90\% of the AWS house and warehouse environments without getting lost and consistently achieves robust localization and high-quality map estimation. 

\subsection{Real robot experiments}\label{sec:realrobots}
We evaluate our system's performance in two different real-world environments over five independent runs each in Table~\ref{tab:combined-results-real} using a SB robot. Experiments were conducted in two room-scale environments—one heavily cluttered and the other mostly clear. Details about our settings can be found on the website. Robot poses are tracked via an Optitrack motion‐capture system to provide high-precision ground truth. 
Robots are equipped with an Intel RealSense D435i RGBD camera, while the onboard IMU supplies high-frequency acceleration and angular-velocity measurements, and the joint encoders provide leg kinematics for proprioceptive odometry. We perform our experiments using these sensors along with the same ROS-based setup as in simulation.
Robots are not equipped with contact estimation sensors, and thus rely on our method as in Section~\ref{sec:contact}.

\begin{figure}[t]
    \centering
    \includegraphics[height=0.33\columnwidth]{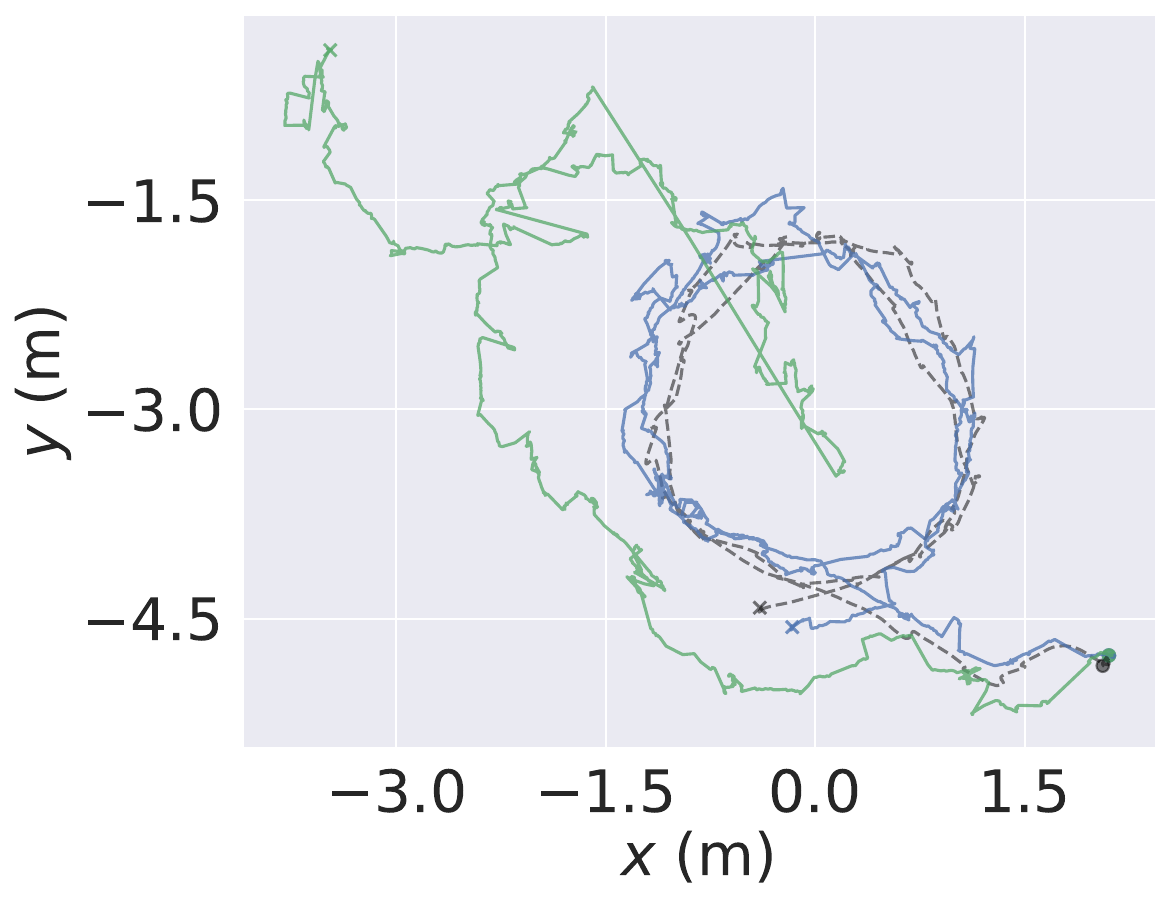}
    \includegraphics[height=0.33\columnwidth]{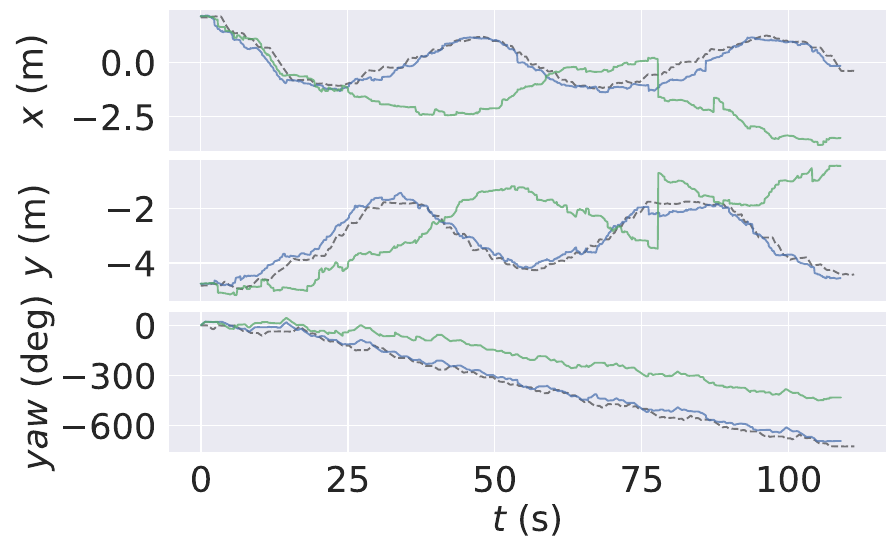}
    \includegraphics[width=0.8\columnwidth]{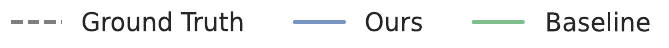}
    \caption{Estimated trajectories using our system and the baseline of the Silver badger robot in a real-world environment (Environment 1, IAS Lab)}
    \label{fig:real-sb-traj}
    \vspace{-1em}
\end{figure}

\begin{figure}[b]
    \centering
    \includegraphics[height=0.3\columnwidth]{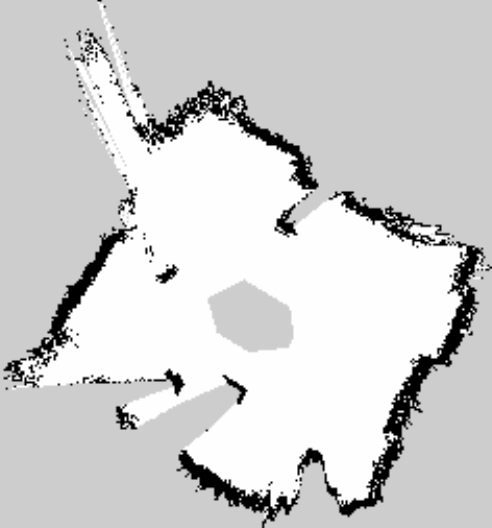}
    \hspace{1em}
    \includegraphics[height=0.3\columnwidth]{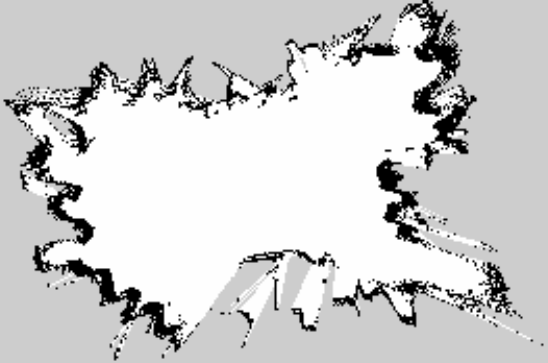}
    \caption{Maps generated on two real-world environments. Left: IAS lab. Right: IRIM lab.}
    \label{fig:real-world-maps}
\end{figure}

Real-world results presented in Table~\ref{tab:combined-results-real} confirm the insights from simulation, showing that our method reduces localization error and generates good-quality maps. The maps for both these environments using our system can be seen in Fig.~\ref{fig:real-world-maps}, showing reasonable performance even in complex cluttered environments such as the IRIM lab. 
Fig.~\ref{fig:real-sb-traj} compares our full system against the baseline for the IAS Lab. Our approach achieves consistently better overall performance and produces a trajectory that follows the ground truth closely. In general, real-world results confirm the need for the full pipeline: leg odometry alone introduces more noise, reducing the accuracy, and only scan stabilization has problems with loop closure, which is improved by the leg odometry edges in the factor graph.  More details about the real-world experiments are available on the project website.

\section{CONCLUSIONS}
This paper shows how to develop a robust navigation system for low-cost quadruped robots, empowering them with the capability of autonomously navigating and exploring the environment.
These platforms are particularly challenging to use due to the limited number of inexpensive sensors available and reactive Reinforcement Learning policies that may cause abrupt movements.
Our system proves that it is possible to make these robots autonomous by using mostly off-the-shelf localization and mapping pipelines, by only fixing some key details.
Our analysis shows that scan stabilization is the key issue preventing the application of standard 2D localization pipelines. Furthermore, we can improve the mapping and the stability of VIO using the information coming from legged odometry.
Both simulated and real-world results prove the effectiveness of our system in terms of map accuracy and autonomy.

In future works, we want to extend this system to more challenging 2.5D/3D mapping, possibly allowing quadruped robots to navigate and explore buildings with multiple floors.





\section*{ACKNOWLEDGMENT}
This project was funded by the National Science Centre, Poland, under the OPUS call in the Weave programme UMO-2021/43/I/ST6/02711, by the German Science Foundation (DFG) under
grant number PE 2315/17-1, and with travel support from the European Union’s Horizon
2020 research and innovation programme under grant agreement No 951847 within ELLIS.


\bibliographystyle{IEEEtran}
\bibliography{bibliography}

\end{document}